\documentclass[conference]{IEEEtran}
\usepackage{cite}

\usepackage{verbatimbox,caption,float}
\newfloat{Listing}
\aptionsetup{Listing}
\usepackage{graphicx} 
\usepackage{svg}
\ifCLASSINFOpdf
\else
\fi
\begin{document}
%
\title{Programming and execution of skill-based human-robot-crane collaborative tasks}

\author{\IEEEauthorblockN{Taneli Lohi}
\IEEEauthorblockA{Intelligent Robotics\\
VTT Technical Research Centre of Finland Ltd\\
Oulu, Finland\\
Email: taneli.lohi@vtt.fi}
\and
\IEEEauthorblockN{Markku Suomalainen}
\IEEEauthorblockA{Intelligent Robotics\\
VTT Technical Research Centre of Finland Ltd\\
Oulu, Finland\\
Email: markku.suomalainen@vtt.fi}
\and
\IEEEauthorblockN{Roope Mellanen}
\IEEEauthorblockA{Konecranes oyj.\\
Hyvinkää, Finland\\
Email: roope.mellanen@konecranes.com}
\and
\IEEEauthorblockN{Tapio Heikkilä}
\IEEEauthorblockA{Intelligent Robotics\\
VTT Technical Research Centre of Finland Ltd\\
Oulu, Finland\\
Email: tapio.heikkila@vtt.fi}
}


%


\maketitle

\begin{abstract}
Highly varying production sets increasing challenges for robotic manufacturing and indoor logistics. New capabilities for agility, flexibility, and robustness are needed. Robot skills, integrating  motions, tool operations, and sensor perceptions consistently provide an execution mechanism for a versatile set of tasks with varying parameters. In this paper, easy-to-use  CAD-model based programming and execution system for parametrized skills and skill monitors is showcased. The execution control structure is dynamic and parametrized, based on a modified Behavior Tree, where only event based communication is used. A human-robot-crane collaborative skill is shown as a test example, where a human instructs an overhead crane and a manipulator in inserting a heavy object supported by the crane, and guided by the manipulator, into the goal. 
\end{abstract}


%
\IEEEpeerreviewmaketitle

\section{INTRODUCTION}

Future manufacturing systems must be provided with capabilities to adapt to highly frequent variations of products (agility), highly frequent changes in product mix and volume (flexibility), and to manage disturbances in material flow and quality (robustness). Robotics plays an increasingly important role in manufacturing automation but faces the challenges to handle high varieties of parts 
whether related to manufacturing operations or indoor logistics operations. 

Sophisticated sensors enable the system to re-adjust and handle a higher variety of products by robots. Sensors also enable the creation and exchange of information within the factory, to recognize and assess situations and integrate the physical and virtual worlds as digital twins \cite{luSmartManufacturingProcess2020}, \cite{kalsoomAdvancesSensorTechnologies2020}. Robot skills, integrating robot motions, tool operations, and sensor perceptions, come to play a key role in providing variability to execute a versatile set of tasks. Easy and fast integration and reuse of available devices and software components become essential.

Robots are not the only moving parts in such production environments. Heavy machines, such as forklifts and overhead cranes, are often vital parts of these production environments. For efficient adaptation of robots and their operation into such environments, robots should efficiently communicate with the "traditional" heavy machinery that operates on industrial PLC. Finally, human operators will be involved in the operations in the foreseeable future, when it comes to high-mix low-volume manufacturing, meaning that the robot programming system must also be able to take the human operators into account. 

Currently, there are no integrated planning and execution SW systems available for developing applications for human-robot-crane collaborative tasks. In this paper, we introduce a complete system for easy management of programming and execution of human-robot-crane collaborative tasks to tackle the challenges of agile and flexible  automation for manufacturing and indoor logistics. The task execution system uses Behavior Tree (BT) -based control flow, which allows parallel skill execution and ability to react to failed skills. We also introduce monitoring skills, which are used to monitor skill execution, detect anomalies, and allow recovery operations. 
The novelties in our approach are the flexibility to configure monitors into skill execution using BTs, and including non-robotic resources, i.e., cranes and human operators, in the system. We include detailed schematics of control and data flow throughout the whole hierarchy of BTs and skill implementation. 
As a test example, we introduce a new human-robot-crane collaborative skill for handling heavy components and parts.  In the next sections, we give a short overview of related work (section 2), present our overall architecture (section 3), principles of skill modeling (section 4), experimental results (section 5), give short discussion (section 6), and finally short conclusions (section 7).

\section{RELATED WORK}
\subsection{Skill based robotics}
There are many definitions for robot skills. Here we generally follow \cite{jiangRobotSkillLearning2024}: robot skills refer to the ability of robots to use their own perception, decision-making, and control capabilities to complete specified tasks. The vast majority of authors agree on a taxonomy for skill based robotics, with layers for tasks, skills, and primitives \cite{pantanoCapabilitybasedFrameworksIndustrial2022}. The robot’s capabilities are first encapsulated in skills, which can then be applied in different tasks by parameterization \cite {heussExtendableFrameworkIntelligent2022} and sequencing. To program an application, the task is specified as a structured flow of skills, after which the skills are parameterized \cite{pedersenRobotSkillsManufacturing2016},\cite{steinmetzRAZERHRIVisual2018},\cite{pantanoCapabilitybasedFrameworksIndustrial2022}. 
According to \cite{bruyninckxSituationalAwareRobotic2025}, the essential activities required to realize robot tasks, expressed as plans, are plan execution, control, perception and monitoring. These activities interact indirectly via (i) information exchange with a world model activity about the current state of the world, and, (ii) decision making delegated at run-time to skill control.
Among the first ones using CAD-based world models was the assembly programming system AUTOPASS \cite{liebermanAUTOPASSAutomaticProgramming1977}. In the AUTOPASS language, assembly specification is refined into robot motion by using a dedicated compiler that composes functional modules. However, the language lacks semantics of CAD primitives and CAD-specific constraints as it was developed before the BREP format became popular \cite{pantanoCapabilitybasedFrameworksIndustrial2022}. There are also intuitive user interfaces proposed for programming such skill-based robots \cite{brunete2016user}. 

There are numerous works that present architectural and programming details for skill-based robotics. The work in \cite{perzyloIntuitiveInstructionIndustrial2016} uses object-centric programming based on geometric constraints between parts to simplify the robot motion programming. The uncertainty-aware robot skills of \cite{thomasNewSkillBased2013a} are generalizable skills that include force-controlled motions but are yet to be unified with the CAD assembly constraints \cite{perzyloIntuitiveInstructionIndustrial2016}. In \cite{paneSystemArchitectureCADBased2020a}, a system architecture and software implementation are presented, facilitating the instantiation of constraint-based skills to execute position and force-based assembly tasks specified with CAD semantics.  The grounding from the CAD-level assembly specification to executable skills involves skill selection and skill parameterization. Skill selection deals with finding suitable skills from the provided library for a given assembly task, by a reasoning module to search through the ontology for the appropriate skills. This has been implemented in Prolog in \cite{perzyloIntuitiveInstructionIndustrial2016}. Once a skill is chosen, its context-dependent parameters are set, and implementation is often performed by constraint-based robot programming. 

SkiROS2, a platform on the Robot Operating System (ROS) \cite{mayrSkiROS2SkillBasedRobot2023}, uses skill formulation based on pre-, hold-, and post-conditions, and automatic programming of skill sequences with a knowledge base and a planner software, similarly to how the work presented in this paper. However, it requires overcoming a steep learning curve in semantic modeling and descriptive programming (cf. PDDL). There are several key differences to the proposed architecture. We provide the user with an easy to use GUI-based planning and programming tool, that use geometric models of work objects to generate parameters for skills, and optionally supporting task programming manually with high-level JSON scripts. We also see the need for monitors to be modeled as parallel skills, instead of tying a single monitor to a single skill, such that monitors can either be used to monitor multiple skills, or multiple monitors with different fallbacks used for single skills. Moreover, SkiROS is not inherently compatible with commercial robot languages, whereas in this paper, we present how a KUKA Quantec KR210  programmed with the KRL (KUKA Programming Language) can be easily interchanged to a commercial crane while keeping the rest of the system intact. Finally, the proposed system can include other heavy machinery besides robots (such as shown with the commercial crane in this paper), as well as with human operators. 



\subsection{Monitoring}
Execution monitoring enables robot systems to prepare for and react to failing executions. Execution monitoring is a continuous real-time task of determining the conditions of a physical system by recording information, and recognizing and indicating anomalies in the behavior. Execution monitoring has been a well studied topic within industrial control, referring to the problem of fault detection and isolation (FDI) \cite{petterssonExecutionMonitoringRobotics2005}.

Execution monitoring can be classified as analytical, data-driven, or knowledge-based \cite{chiangFaultDetectionDiagnosis2012}. Analytical approaches rely on the concept analytical redundancy and analytically generated quantities, obtained from different sets of variables are compared. In data-driven approaches, the information used for monitoring is derived directly from input data. The decision making is often based on statistical methods.  

Knowledge-based approaches are generally designed to simulate the problem-solving behavior of human experts. Both analytical and data-driven approaches apply to knowledge-based approaches, which can be divided into causal analysis, expert systems, and artificial neural networks \cite{petterssonExecutionMonitoringRobotics2005}. Causal analysis methods are based on causal modeling of fault-symptom relationships, e.g., a signed directed graph (SDG) \cite{petterssonExecutionMonitoringRobotics2005}, reflecting the behavior of the equipment involved as well as the general system topology.

Combining the skill based programming with an architecture based on skills as finite state machines is given by \cite{herreroSkillBasedRobot2017}. Monitoring operations are introduced also as a skills, and assembly, vision and Workspace Monitoring skills are integrated into a workflow. The Workspace Monitoring skill is continuously supervising the environment, allows tracking a human position in the workspace, and interacts with Cartesian/articular motion state in order to control or stop motions. Because skills are independent modules, they interact with each other using a Parameter Server, which is essentially a ROS namespace.

In \cite{riosMethodologyRobustlyMonitoring2023}, a Finite State Machine (FSM) is used to represent skill control for contact rich tasks, and a methodology is given to define monitors based on a contact graph. Monitors are used to trigger transitions between the states, and as such, the monitors are used for discrete time sensor feedback rather than anomaly detection, as in our case.

An architectural approach for integrating monitors for work flow control is given in \cite{vanderseypen2025online}. The goal is to be able to autonomously adapt control and perception behavior to the overall situation, using knowledge graph about which control and perception algorithms are available at any instant and select the appropriate ones. A skill selection activity queries the knowledge graph at runtime, to find which skill configurations can be used for a particular task in a particular environment, and a skill execution activity reconfigures and coordinates accordingly the running activities inside the robot’s software architecture.

\subsection{Behavior Trees}
BTs provide a task-switching scheme for robots to adapt to unforeseen situations, recover from errors, and for for operating effectively and securely in dynamic and unpredictable environments \cite{gugliermoEvaluatingBehaviorTrees2024}. Using BTs to organize skill based control in robotic tasks offers a dynamic control structure that allows to construct and restructure program at run-time. Advantage over a finite state machine (FSM) is that individual BT nodes are reusable components, that doesn't depend on other nodes, when in FSM, transitions must be redefined when used in a new context  \cite{bagnellIntegratedSystemAutonomous2012}. 

BT's have been applied in varying but rather simple applications to introduce on-line adaptability, like in \cite{hutter-mironovovaBehaviorTreeDecision2025} for online selection of grippers. In \cite{wuMicROSBTEventDrivenBehavior2021a} event based BT  ticks tree, when any node return Success or Failure, or when blackboard (shared memory that all nodes have access to, used for communication between nodes \cite{colledanchiseImplementationBehaviorTrees2021a}) is updated. This is made to save CPU resources when nothing happens. Ticking is done to check if any condition that effect capability to perform the current action has changed (e.g., in mobile robotics, battery level is low, or in industrial robotics, someone has entered in a robot workspace). The same behavior, as we do, can be achieved using monitors parallel to the relevant sub-tree. For robustness, the design of the BT should aim to increase its reactivity. This is because a robust BT is one that remains correct in the presence of environmental changes, and increased reactivity allows the BT to quickly adapt to such changes \cite{gugliermoEvaluatingBehaviorTrees2024}.

\section{ARCHITECTURE}

\subsection{Three layer architecture}

\begin{figure}[htp]
      \centering
      \includegraphics[scale=0.20]{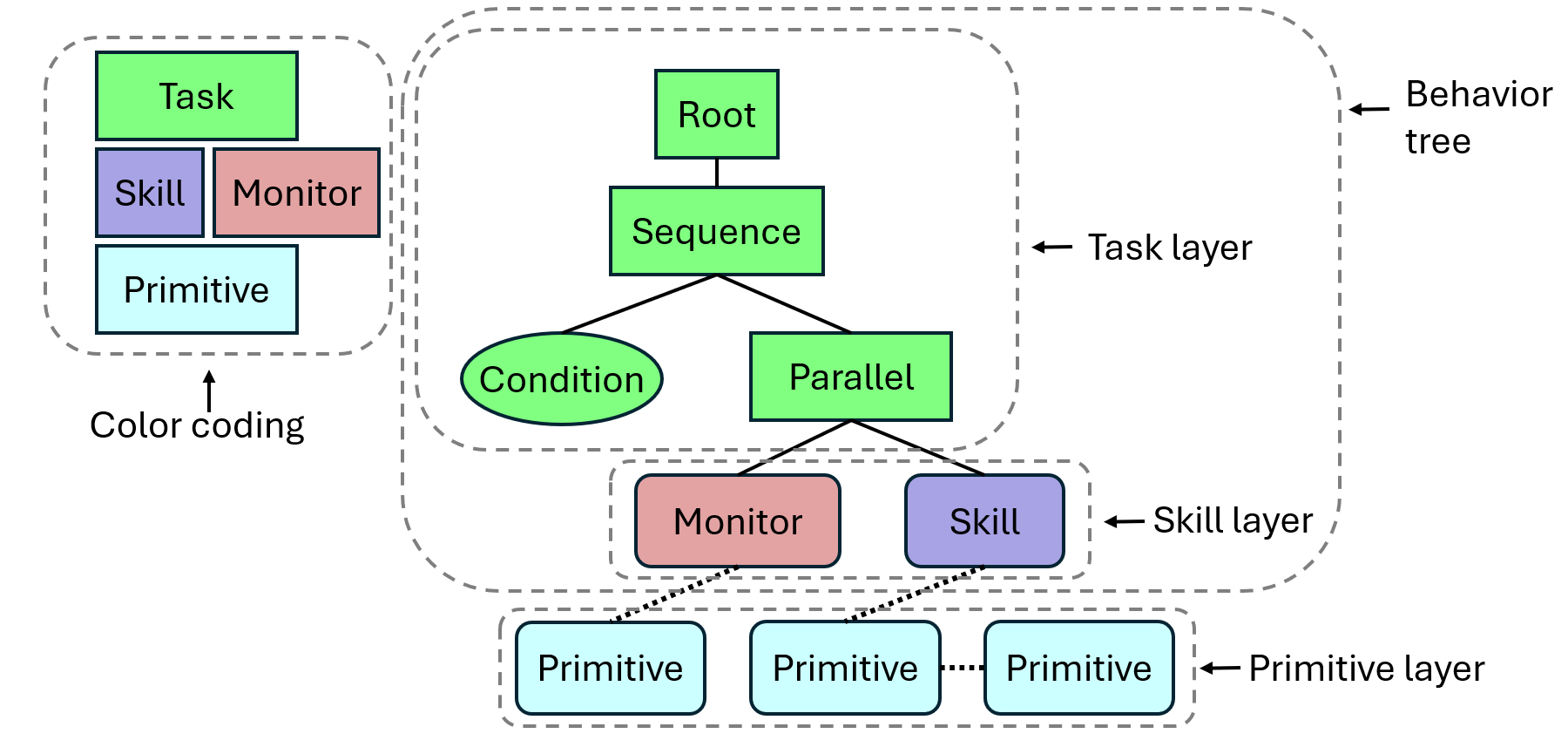}
      \caption{Three control layers of the robot control system. Task layer (green) contains behavior tree control nodes and condition nodes, Skill layer (red and purple) contains skills and skill monitors that are leaf nodes in the behavior tree, and Primitive layer (cyan) contains primitive operations that skills use to interact with the environment.  }
      \label{TreeLayerArchitechture}
   \end{figure}

Our control architecture for skill-based robot control follows the principles of the common three-layer architecture with layers for tasks, skills, and primitive operations \cite{pedersenRobotSkillsManufacturing2016}, \cite{pantanoCapabilitybasedFrameworksIndustrial2022} (Figure \ref{TreeLayerArchitechture}). Parametrized skills and primitives are implemented and tested components in a skill library \cite{heikkilaManufacturingOperationsServices2022}. Our approach compares to the one by \cite{paneSystemArchitectureCADBased2020a}, but with CAD-based sequencing of skills to a task from the skill library by a human operator. Tasks don’t have predefined control structures, but they are created by an operator or programmer using a modified Behavior Tree, for which skills are selected from the skill library. A task is planned by creating a control structure using sequence nodes, fallback nodes, and modified parallel nodes from the modified Behavior Tree, where skills and condition nodes are leaf nodes in the tree. The task is created, and skills are parametrized using a graphical user interface where parameters are either derived from the CAD model of the work object, or the operator/programmer fills the required parameters manually. A task control recipe in JSON format with control structure and parametrized skills is generated and forwarded to a robot control system. The robot control system, equipped with cell configuration data, executes the task according to the control recipe. 

\begin{figure}[htp]
      \centering
      \includegraphics[scale=0.3]{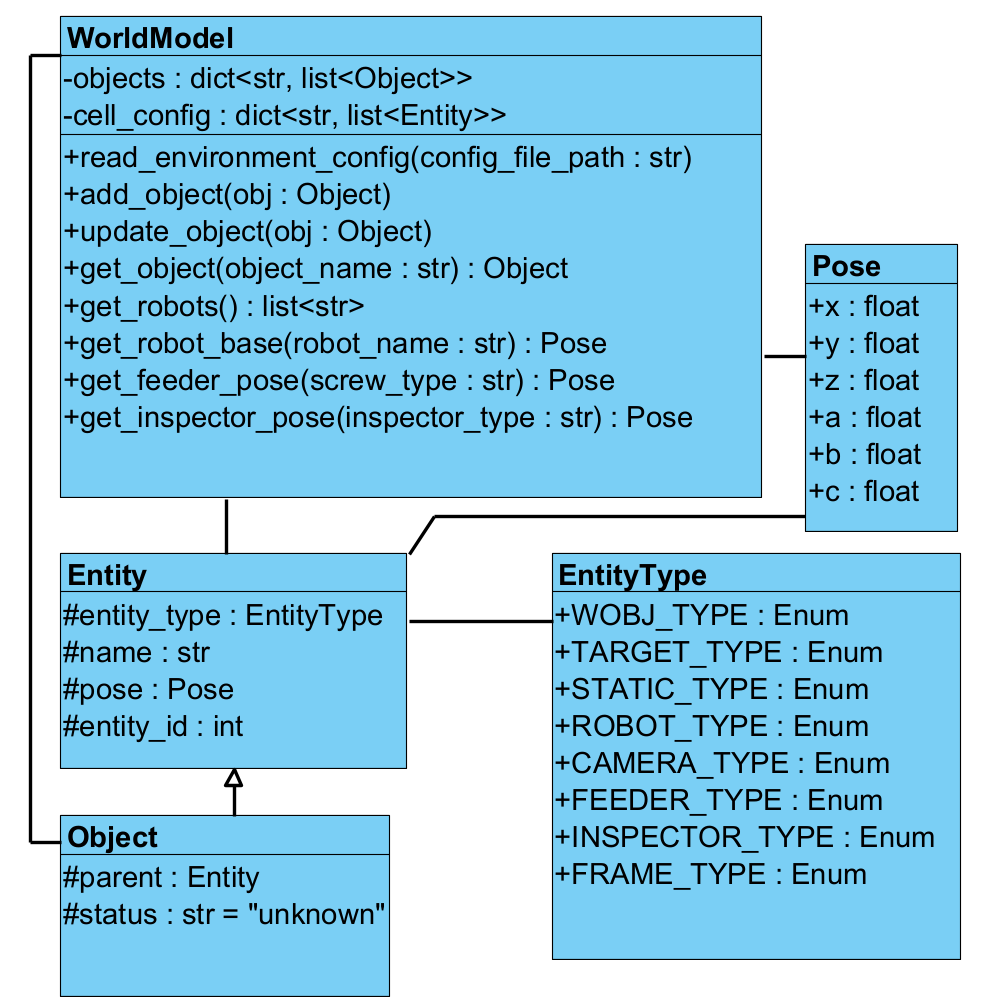}
      \caption{World model structure. The world model contains runtime information about objects in the robots workspace and cell configuration information. Information from model is queried using \textit{get} methods. Perception primitives updates world model with object poses. Entity is a base class of things in the world, which is derived to add more attributes that different entity types require.}
      \label{WorldModel}
   \end{figure}

Skills and primitives have a predefined control sequence that includes synchronization between skills and primitives. Skills are parametrized, and some parameter values are provided by the control recipe, and some are queried from a world model (Figure \ref{WorldModel}) during skill execution. Skills interact with the physical world using primitives.  Primitives are atomic components that are used to move a robot, actuate tools, or trigger perception operations. Primitives can be synchronous, like on-line planning an approach path for grasp, or asynchronous, like executing a grasp using a robot.  Perception primitives update also the world model, typically with poses of objects. Poses of devices and tools, like robots, tool magazines, feeders, and cameras are given in the cell configuration part of the world model. Monitor skills are special types of skills that do not directly interact with the physical environment but monitor the execution of skills to find anomalies in task execution. Monitors don’t typically trigger primitives but use data that primitive operations produce and communicates detected failures to task layer to enable recovery skills.

\subsection{Modified Behavior Tree}
We are using the Behavior Tree (BT) as a control structure to create an execution order for skills in a task. A BT is a tree of nodes and leafs, where the first node is a root node that has one child. Other control nodes can have multiple children. The other nodes are control flow nodes (sequence, parallel or fallback), or leaf nodes (execution nodes or condition nodes). The child nodes may consist of other control nodes, or, if they are at the leaf level, in our case condition nodes or skill and monitor nodes as execution nodes. Our approach to apply a BT is illustrated in Figure \ref{TreeLayerArchitechture}. 

Our version of BT differs from the standard implementation \cite{PyTreesPy_trees}, \cite{colledanchiseImplementationBehaviorTrees2021a} in two ways. Instead of tick-based execution, our version is event-based. We also modified the logic that parallel node uses to report its status to react instantly to \textit{failure} events. In this way we introduce more reactive overall behavior.

When a modified parallel node receives a \textit{start} event, it will send a \textit{start} event for each child. The children run simultaneously until \textit{M} children sends a success event or \textit{K} children sends \textit{failure} event, where \textit{M} and \textit{K} are user defined thresholds. Then the node sends a \textit{success} or a \textit{failure} event to its parent, send \textit{shutdown} event to other active children, waits for children to shut down and then shuts down itself. Difference to standard BT parallel node is that in our version we can define how many \textit{failure} events are needed before \textit{failure} event is reported to the parent. That allows having multiple monitors parallel to a skill and if any monitor sends \textit{failure} event, \textit{failure} is reported to parent. Because monitors only return value to parent if a \textit{failure} is detected, and never return \textit{success}, changing \textit{M} gives opportunity to also have multiple skills parallel, and the node will send \textit{success} only when \textit{M} skills is successfully completed, while monitors are running parallel.  In standard version there is only one constant \textit{M}: how many children \textit{success} is needed to report \textit{success} to parent. Required \textit{failures} depends on \textit{N} - \textit{M}, where \textit{N} is number of children \cite{colledanchiseBehaviorTreesRobotics2018}.

\section{MODELING AND IMPLEMENTATION OF TASKS AND SKILLS}

Control flow and synchronization of skills, monitors and primitives are modeled using UML activity models. Each “swim lane” in the activity model diagram represents a lifeline of a control sequence, which can be either a skill or a primitive. 

Figure \ref{TreeLayerArchitechture} shows the three layers of our skill based control. Communication between the layers is illustrated in the activity diagram (Figure \ref{GraspLoadActivity}). In designing the skill models, the task-layer is not modeled, and in Figure \ref{GraspLoadActivity} it is included for illustration only.  Task activity (green) contains the world model (Figure  \ref{WorldModel}) and a Behavior Tree (control and condition nodes). The Behavior Tree can make precondition checks (condition node, i.e. if an object is localized) from the world model before starting skills. Figure \ref{GraspLoadActivity} illustrates how a skill communicates with the task layer, by querying object poses from the world model and with the robot controller by activating a primitive operation, and how a monitor uses primitives as a data source. Primitive operations can be run as actions inside the skill activity (Plan trajectory) or can be executed as an own activity in a separate life-line, and execution is triggered by the skill (Grasp). 

\subsection{CAD based planning of task and skills}

\begin{figure}
    \centering
    \includegraphics[width=1\linewidth]{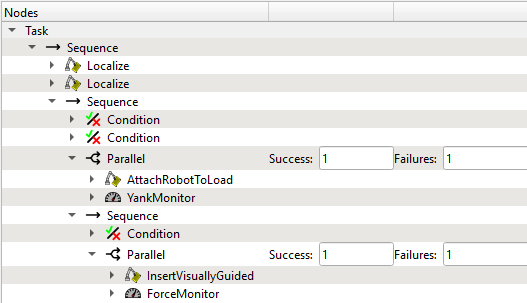}
    \caption{Task created using GUI. Little robot arms are skills, gauge symbol represent monitor, arrows are Behavior Tree control nodes and Conditions are Behavior Tree condition nodes}
    \label{TaskGUI}
\end{figure}

A CAD based task planning and programming system has been developed using the OpenCascade Technology CAD kernel \cite{OpenCASCADETechnology}. The task planner SW includes visualization of a part or assembly CAD model. The model is interactive, and features like faces, edges and vertices can be selected from the model. A task is created by building a control tree using BT control nodes and skills (Figure \ref{TaskGUI}). Available skills and monitors that are included in the skill library are selected from a list, connected to objects and added as leaf nodes in the tree. Compatibility checks between skills is supported by comparing pre- and postconditions from skill description.

The skills are parametrized one-by-one by selecting features from the objects in CAD model as they are connected to the skills. E.g., a pose can be defined using a point and two perpendicular edges, or data fields can be filled by the operator (Figure \ref{parametrizedSkills}). Available control nodes, skills and monitors are displayed in Table \ref{SkillsAndMonitorTable}.

\begin{table}[h]
\caption{Control nodes, skills and monitors available for use}
\label{table_example}
\begin{center}
\begin{tabular}{|c||l|}
\hline
Control nodes & 
 \begin{tabular}{@{}l@{}}
    Sequence \\
    Parallel \\
    Fallback 
  \end{tabular} \\
\hline
Condition nodes & 
 \begin{tabular}{@{}l@{}}
    Object localized
  \end{tabular} \\
\hline
 Skills & 
 \begin{tabular}{@{}l@{}}
    Localize \\
    Pick \\
    Place \\
    Attach robot to load \\
    Collaborative move load \\
    Insert visually guided \\
    Screw 
  \end{tabular} \\
\hline
 Monitors & 
 \begin{tabular}{@{}l@{}}
    TimerMonitor \\
    ForceMonitor \\
    YankMonitor 
  \end{tabular} \\
\hline
\end{tabular}
\end{center}
\label{SkillsAndMonitorTable}
\end{table}

\subsection{Implementation of tasks, skills and primitive operations}
The core of our skill-based robot control system is the task controller SW component. The task controller is implemented in Python as a ROS2 node and uses the skill-library to create and execute a task, according to a task control recipe. The control recipe is implemented in JSON format, which is passed from planning software to robot control system as a file or using a TCP/IP connection. The task controller reads the recipe and creates a BT based control tree, with parametrized skills, based on the recipe. Robot cell configuration data is embedded in the configuration files of HW and SW components. Runtime environment data is stored in a world model (Figure \ref{WorldModel}), which is implemented as a python class, inside the root node, and from which skill executors can query information at runtime. The world model contains information about localized objects in robot workspace and it is populated by Localize -skill and updated by handling skills. 

\begin{figure}[h]
    \centering
    \includegraphics[width=1\linewidth]{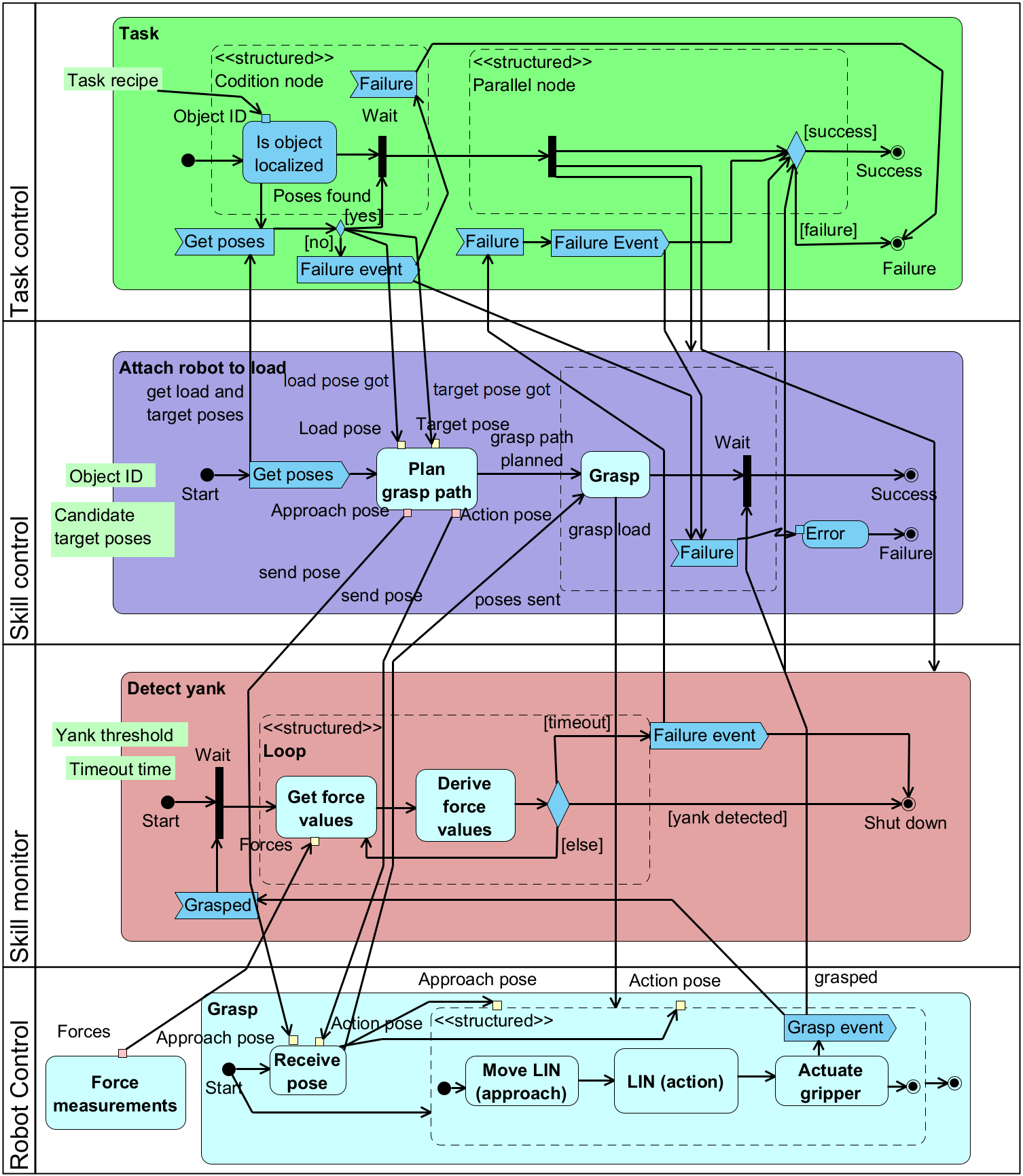}
    \caption{An example of a control flow in “Attach robot to load” -skill, its primitive operation "Grasp" and a parallel monitor "Detect yank" that detects if the grasp is successful. (purple=skill, red=monitor, cyan=primitive operation) 
    }
    \label{GraspLoadActivity}
\end{figure}

\begin{figure}[h]
    \centering
    \includegraphics[scale=0.35]{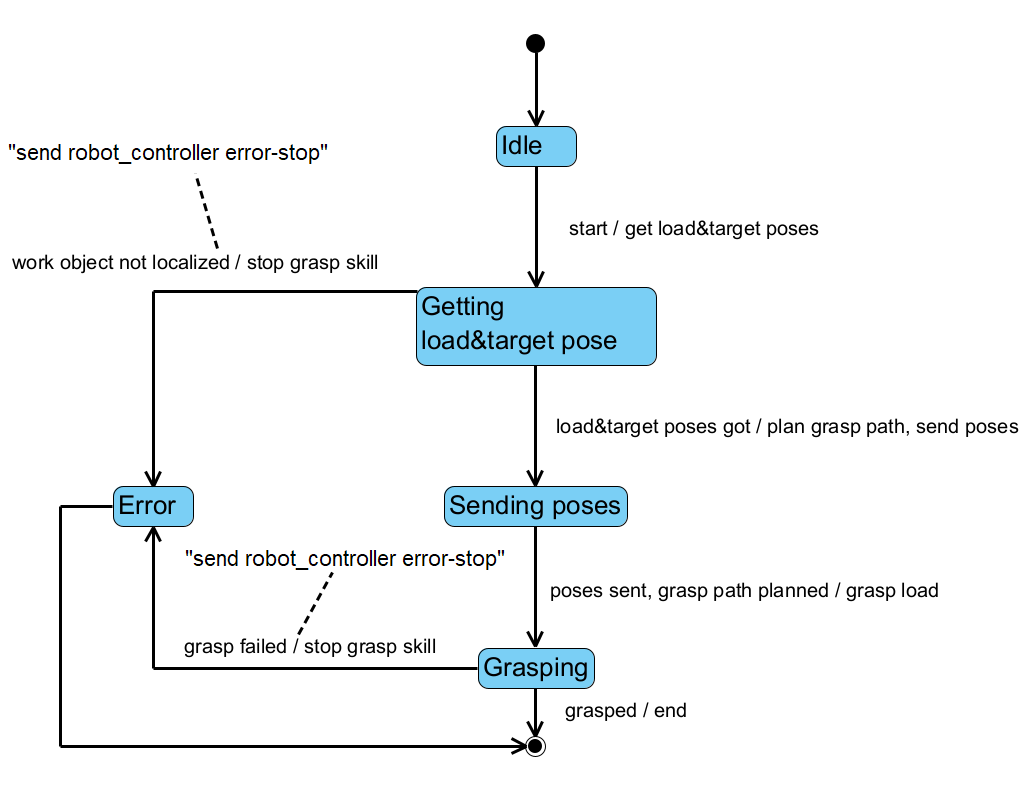}
    \caption{State-transition diagram of "attach robot to load" skill.}
    \label{GraspLoadStateTrans}
\end{figure}

Skills have a predefined control sequence that is designed using UML activity models and implemented using a FSM. Each action in activity diagrams (Figure \ref{GraspLoadActivity}) corresponds to transitions in the state-transition diagram (Figure \ref{GraspLoadStateTrans}).  Skill executor is a module that is run separately and is responsible for executing the skill. Each skill executor has a communication interface that is used to send events, messages and queries to parent nodes in BT or to a root node that manages the task. Synchronization to primitive operations is implemented using python functions (synchronous) or ROS2 Services and Topics (synchronous and asynchronous).
 
Primitive operations are implemented using python functions, C++ executables or as program blocks in robot language programs (e.g. KRL, KUKA Sunrise, URScript, or RAPID languages). Asynchronous primitive operations have ROS2 service interface to start operation and use ROS2 Topics to communicate about the state of the operation.

\begin{figure}[h]
    \centering
    \includegraphics[width=1\linewidth]{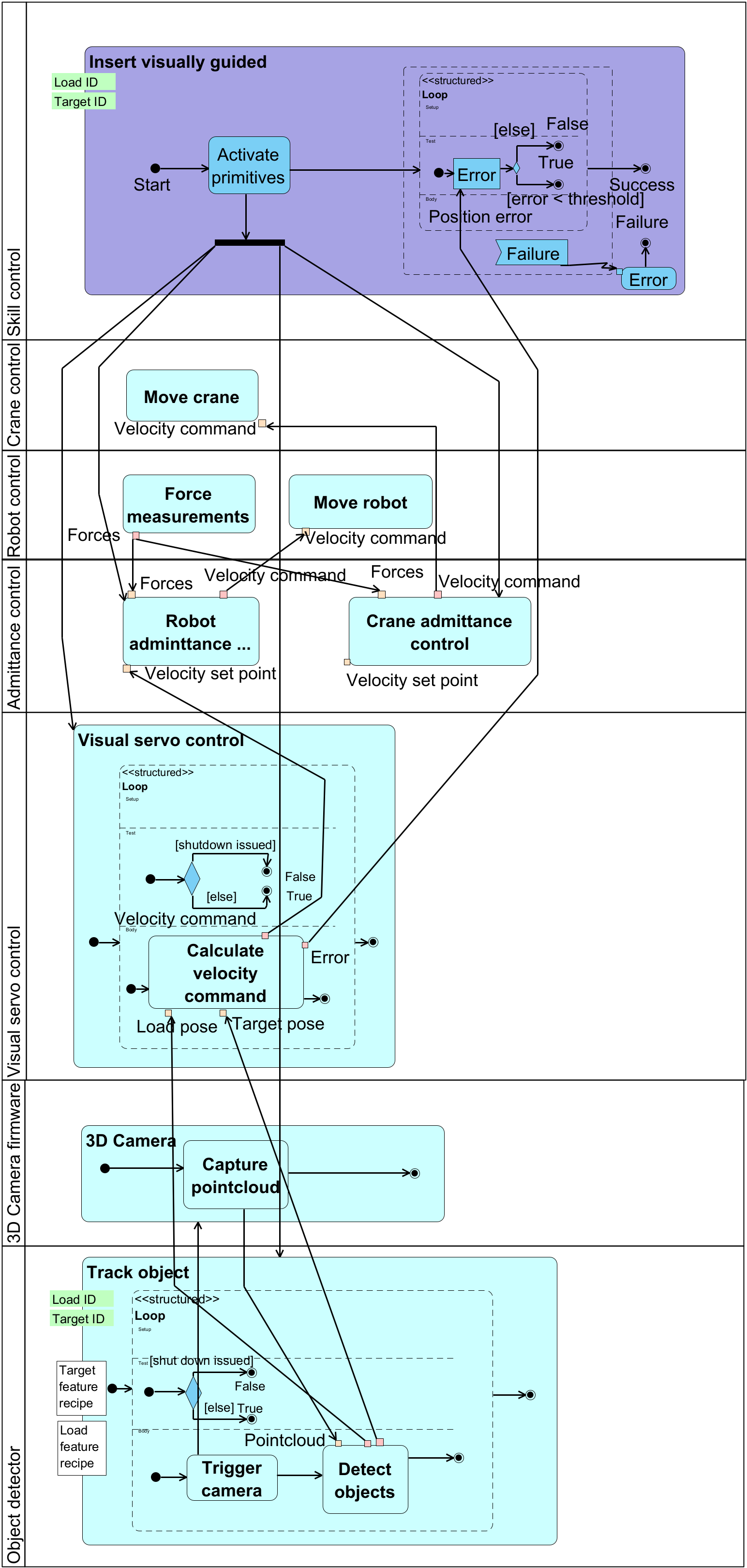}
    \caption{Control flow and synchronization of "Insert visually guided" skill and its primitives. (purple=skill, cyan=primitive operation) }
    \label{InsertVisuallyActivity}
\end{figure}

\begin{figure}[h]
    \centering
    \includegraphics[scale=0.25]{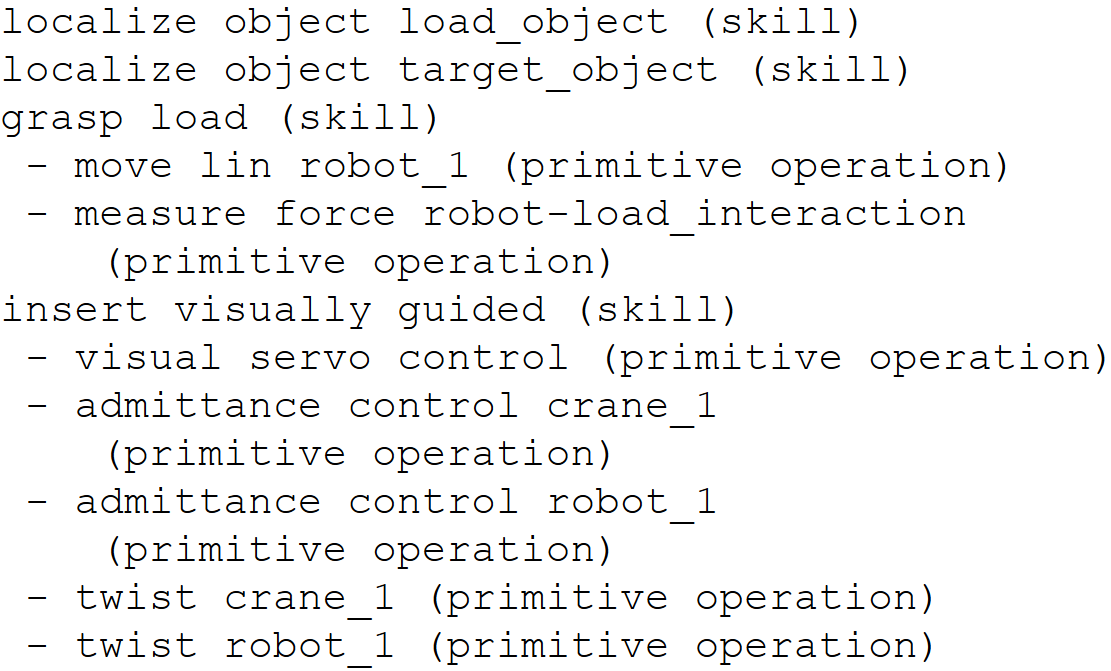}
    \caption{Task scenario: sequence of skills and primitive operations }
    \label{TaskScenario}
\end{figure}
\section{EXPERIMENTS}

\subsection{Human-robot-crane control}

\begin{figure}
    \centering
    \includegraphics[width=1\linewidth]{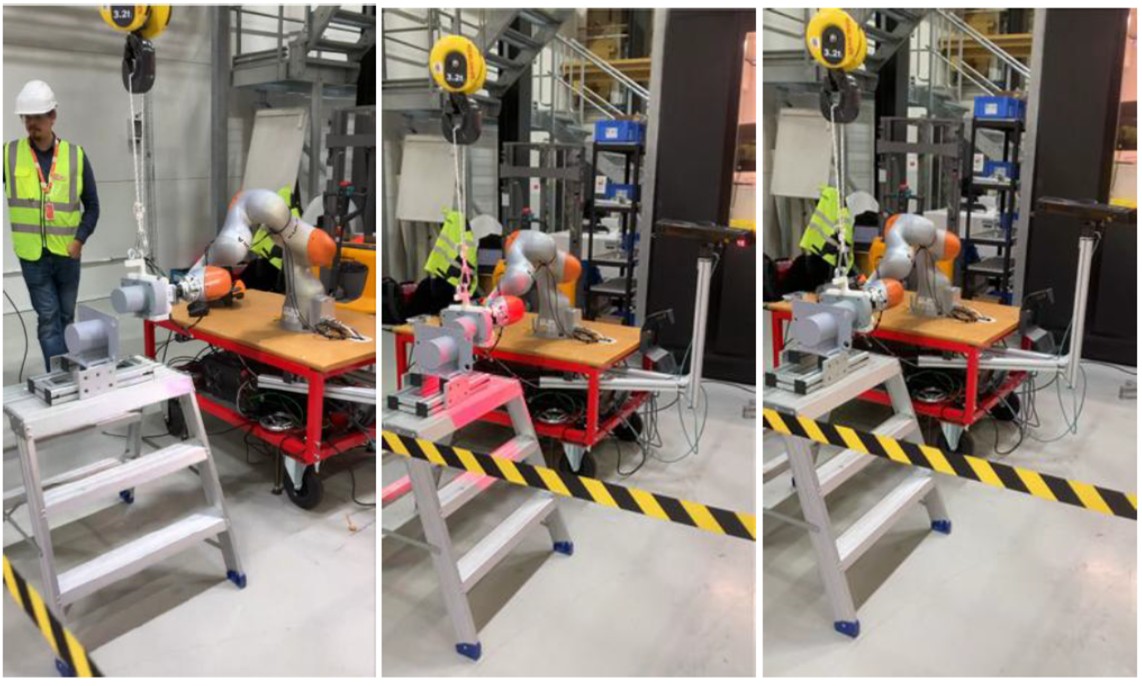}
    \caption{Test sequence: the load is carried by the overhead crane, the robot pushes the load to an attachment position. After this the operator attaches the load with screws.}
    \label{RobotOperatorCollab}
\end{figure}

We have tested our programming and control system in an assembly task with human-robot-crane collaboration. The goal is to use collaborative control of an overhead crane and a light-weight robot to move a heavy load to a target object, after which a human operator attaches the load to the target using screws (Figure \ref{TaskScenario}, Figure \ref{RobotOperatorCollab}). Motivation to use robot-crane collaboration is to remove the human operator from having to push the heavy load with one hand while simultaneously controlling the crane through remote controller with the other hand.  In this experiment, we used 3D printed mockup parts as a load and a target object, but the same principle works on heavier objects as well. 

Both the robot and the crane motions are controlled with admittance controllers, transforming the contact force from the robot against the load to velocity control signals of the robot, and the crane \cite{rosalesRobotOverheadCrane2025}. Target motion references for the load are generated either using joystick control by the human operator, or a 3D visual servo controller \cite{10665161}, where an object detector equipped with a 3D camera is used to track 6D poses of the load and the target.    

\begin{figure}
    \centering
    \includegraphics[width=1\linewidth]{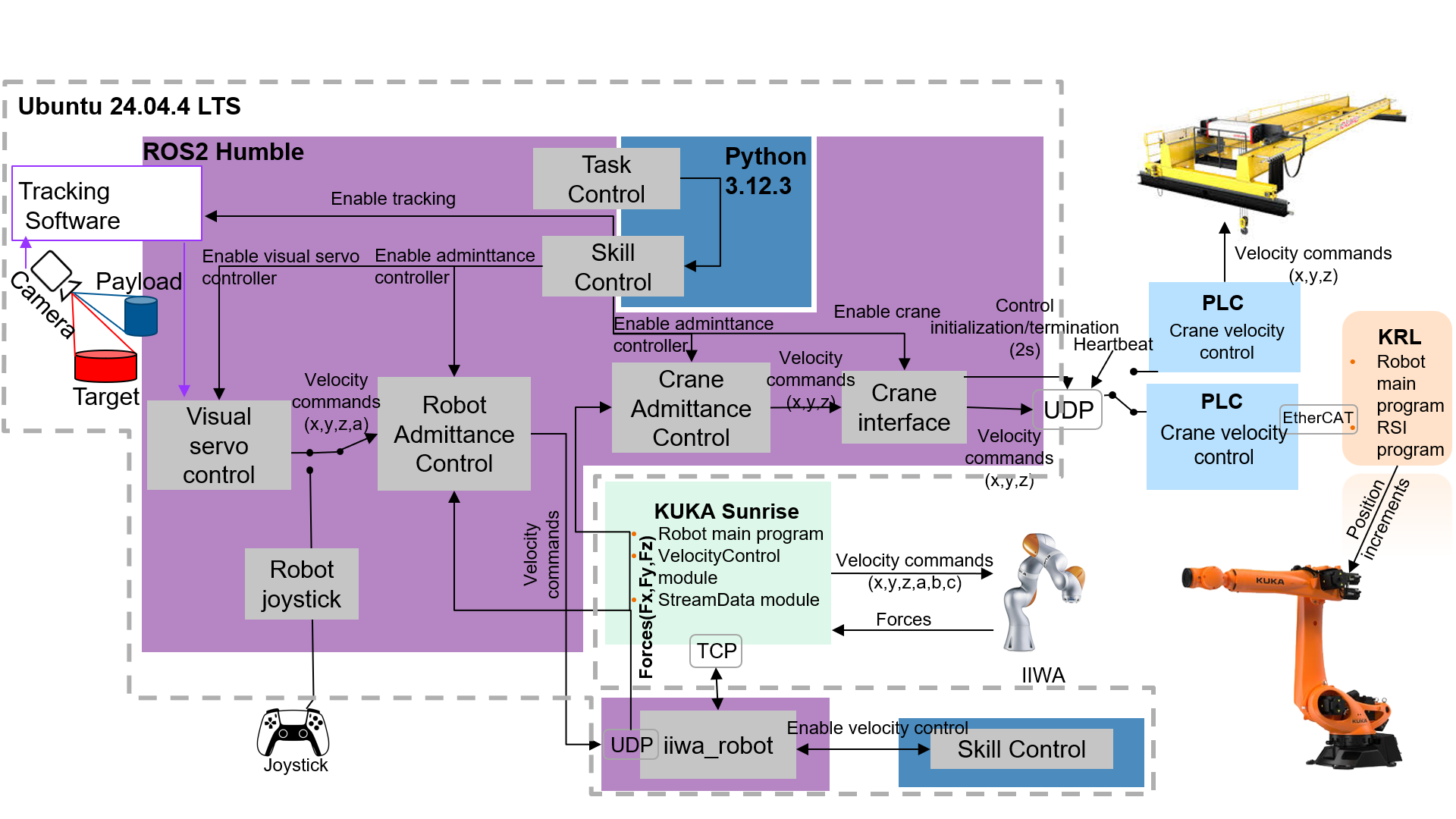}
    \caption{HW/SW implementation architecture -  cranes: Konecranes CXT Smart crane, KUKA Quantec KR210 R2700-2 as a crane emulator, robot: KUKA LBR iiwa 14 R820, equipped with a magnetic gripper}
    \label{SW_HW_ArchitectureDiagram}
\end{figure}

In our experiment, the test crane was a Konecranes CXT Smart crane, controlled with Beckhoff’s PLC, and with a maximum load of 3200 kg. Alternatively, we have used a KUKA Quantec KR 210 as a crane emulator, with which the robot-crane collaborative skills were first tested. The set-up for the real overhead crane was done easily in a plug-and-play manner. The collaborative robot system is a KUKA LBR iiwa14 robot arm and Sunrise controller, Schmaltz SGM-HD-S 70 magnetic gripper, and Photoneo MotionCam 3D camera. The 3D camera is calibrated to the iiwas base coordinate frame with proprietary computer vision software and it is used to track the load and the target. ROS2 Services and Topics are used to communicate between the SW components (Figure \ref{SW_HW_ArchitectureDiagram}). The overall SW/HW architecture is shown in Figure \ref{SW_HW_ArchitectureDiagram}. 

\begin{figure}
    \centering
    \includegraphics[width=1\linewidth]{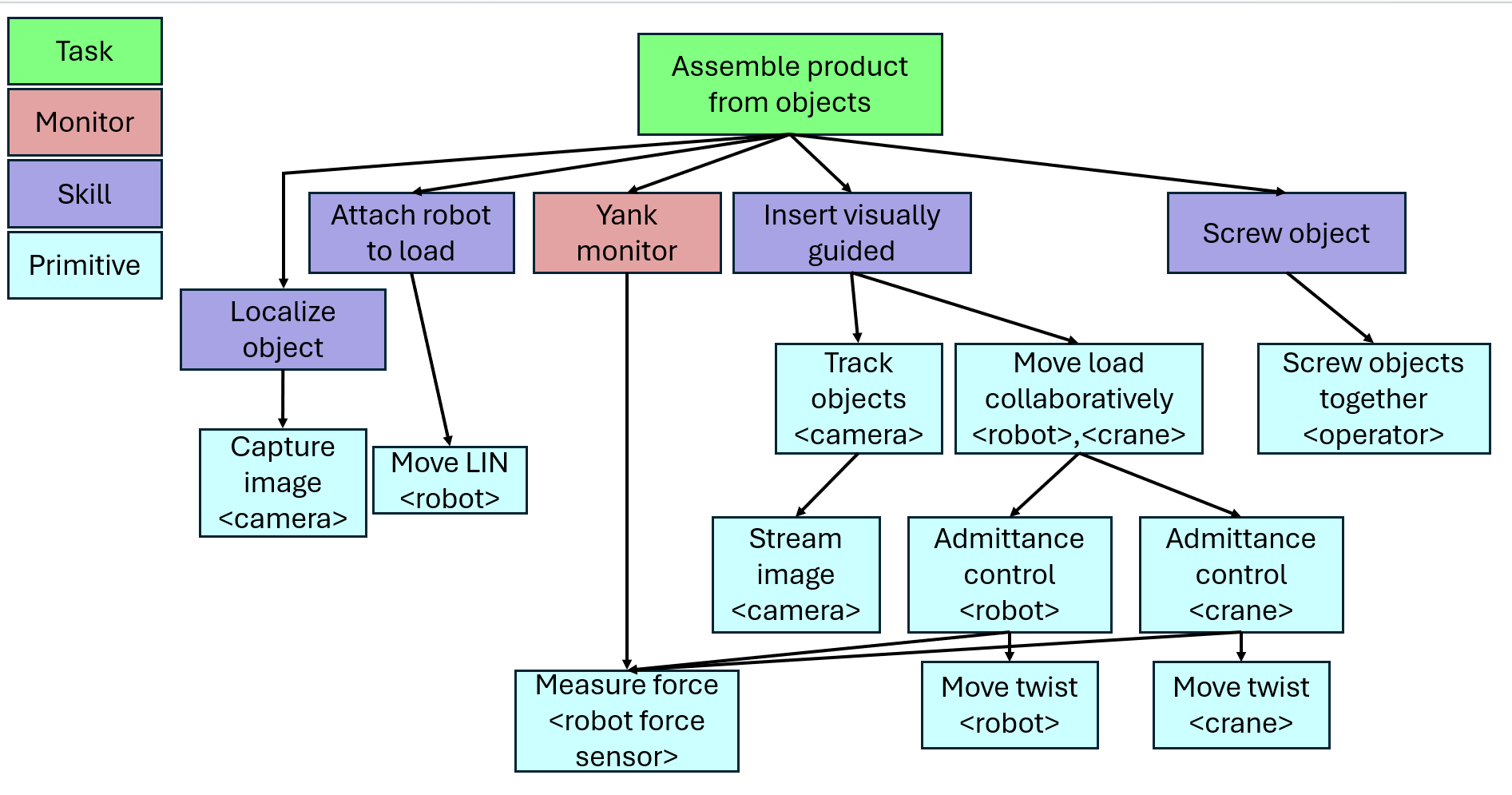}
    \caption{Control hierarchy in human-robot-crane collaboration control: task, skills, monitors and primitive operations used in the assembly task.}
    \label{SkillsAndPrimitivesInTask}
\end{figure}

\begin{figure}
    \centering
    \includegraphics[width=1\linewidth]{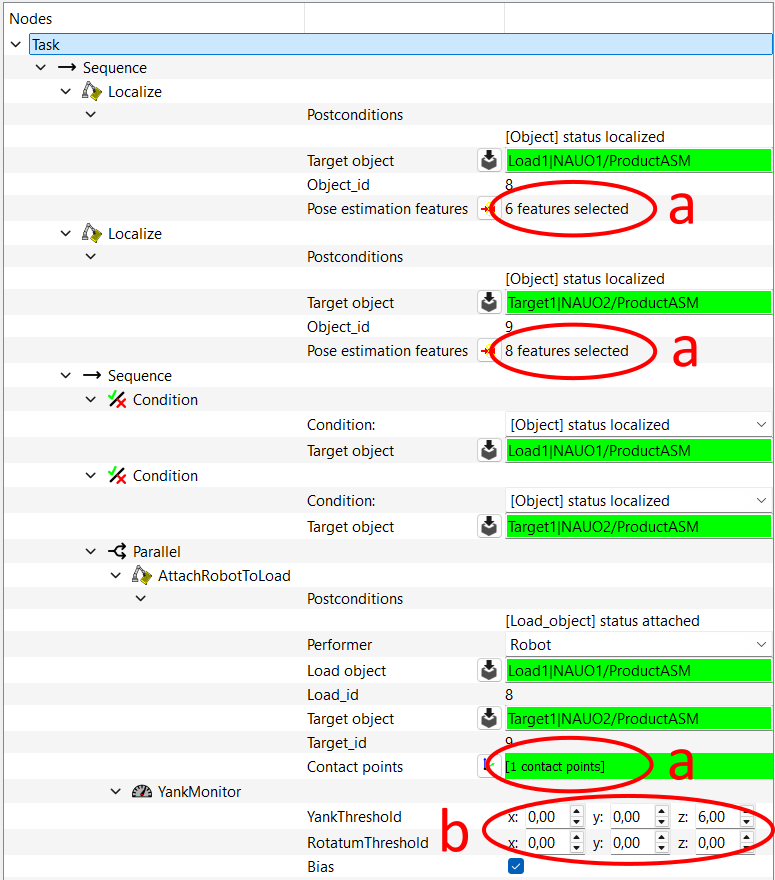}
    \caption{Parameters for skills are either generated using CAD model of object (a), or given by programmer (b).}
    \label{parametrizedSkills}
\end{figure}

\begin{figure}
    \centering
    \includegraphics[width=1\linewidth]{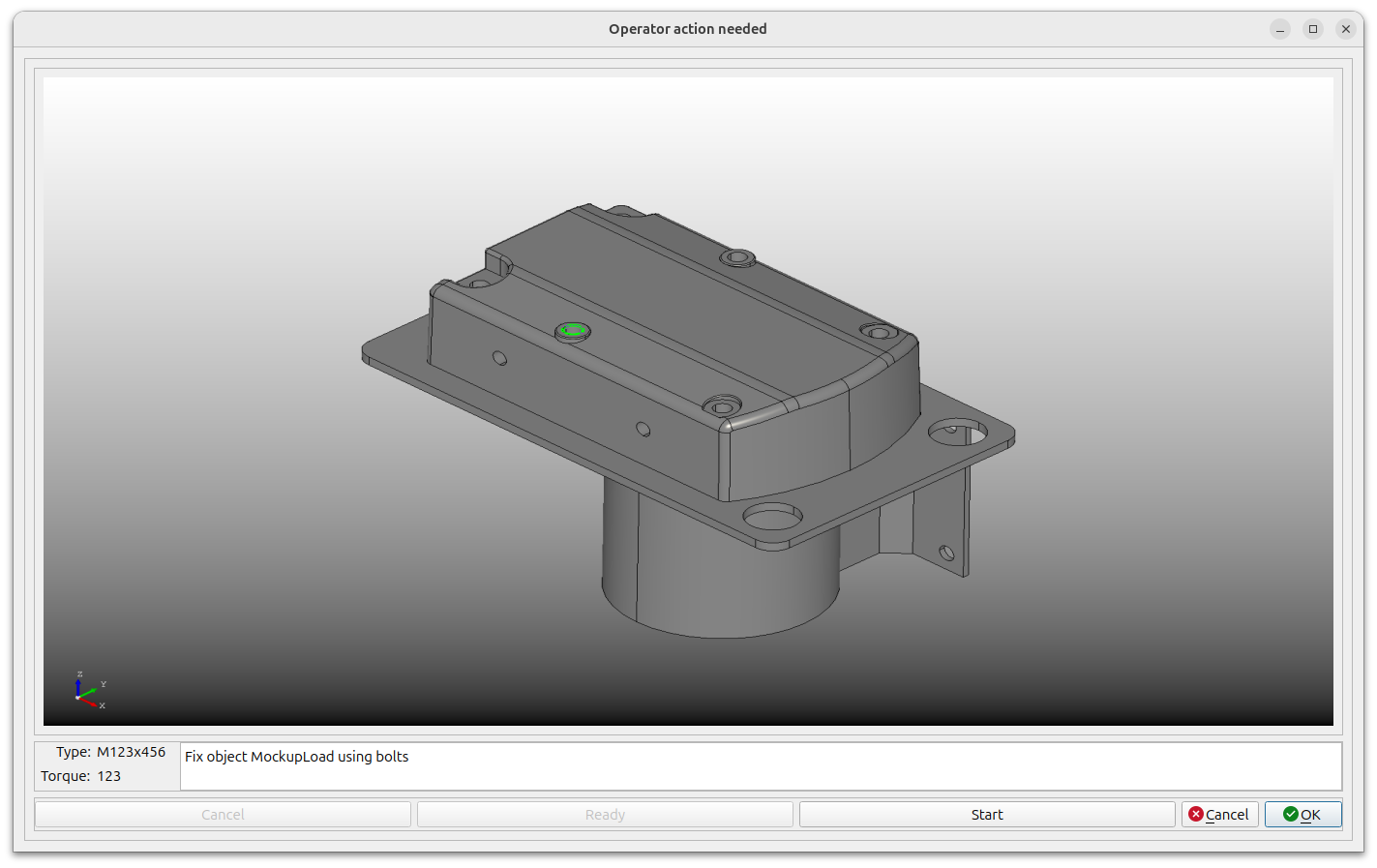}
    \caption{Screw prompt for operator, green highlighted screw hole displays target location for screw. Screw type and target torque is given in bottom left corner. }
    \label{OperatorScrewUI}
\end{figure}

The task and related skills, monitors and primitive operations used in the test are illustrated in Figure \ref{SkillsAndPrimitivesInTask}. The tested task was planned as a skill sequence and skills parameterized by a programmer using our programming tool (see Figures  \ref{TaskGUI} \& \ref{parametrizedSkills}). A  CAD model of the assembly was used to generate the parameters for the skills (e.g. grasping pose, waypoints for visual servoing and screw locations). Other needed parameters the programmer entered manually (e.g. threshold for a Yank monitor, screw types and target torques). The parametrized task recipe in JSON format was given to the task controller, in which the task executor generated the BT for the task, and started executing it. First the load and the target were lozalized by the \textit{Localize object} skill, based on which the \textit{Attach robot to load} skill planned the local path for grasping and grasped the load with Iiwa and its magnetic gripper. Parallel to the \textit{Attach robot to load} skill the \textit{Detect yank} monitor was running (yank is first time derivative of force) to detect if grasp is successful. If it is not the monitor will report a \textit{failure} and the task is halted. After succesful grasp, the  \textit{InsertVisuallyGuided} skill (see Figure \ref{InsertVisuallyActivity}) used the admittance controllers of the robot and the crane and visual servo controller  to bring the load near the target. Then \textit{Screw object} skill prompted the human  operator (Figure \ref{OperatorScrewUI}), which type of screw belongs to which screw hole and the human operator performed screwing and reported success using the UI (Figure \ref{OperatorScrewUI}). 

\section{DISCUSSION}

The use of BT with parallel nodes allows running a monitor or even several monitors parallel to a skill, to ensure the skill is executed properly (a difference to SkiRos2, where the monitors are inside the skill module). The system is built to allow the use of fallback nodes to enable corrective operation after detecting anomalies, where the ability to have multiple monitors, or monitors monitoring multiple parallel skills, becomes essential. Different fallback options are, for example,  inverting skills, activating alternative skills, or leaving the decision to an operator. Decorator nodes that allow custom behavior, such as loops, are also planned to be included. 


For skill implementations, we have followed the traditional approach \cite{jiangRobotSkillLearning2024}, where robot skills are mostly implemented according to manually defined programs. Control flows for skills are specified with UML Activity models and mapped to easily programmable state-transition models, illustrating the behavior and collaboration of the skills and related primitive operations. Robustness is introduced by precondition checks and execution monitors linked to skills, which are also highlighted to a task programmer in our flexible task programming SW. To adapt to uncertainties and inaccuracies in the operating environments, perception operations are tightly integrated with tasks as perception skills. Programming task sequences in the form of recipe scripts with skill and monitor parameter settings is made easy and fast using CAD model data. Finally, as the system is agnostic to the origin of the skills, machine learning methods can be applied for learning the parameters, or even for introducing completely new skills or primitive operations to the system.  


\section{CONCLUSIONS}
We have introduced a system for easy programming and flexible execution of robot tasks to tackle the challenges of agile and flexible robot automation for manufacturing and indoor logistics. Our solution is based on modeling robot activities as skills, integrating perceptions and robot operations to purposeful workflow sequences. Flexibility and robustness are introduced by BT based control over skills, with programmable monitors.

\section*{Acknowledgment}
This work was supported by the INVERSE project, funded by the European Union, Horizon Europe research and innovation programme (Grant Agreement 101136067).

\bibliographystyle{IEEEtran}
\bibliography{IEEEabrv, refs}



%



\end{document}